\documentclass[letterpaper, preprint, paper,11pt]{AAS}	

\usepackage{bm}
\usepackage{amsmath,amssymb}
\usepackage{subfigure}
\usepackage{algorithm}
\usepackage{svg}
\usepackage{dirtytalk}

\usepackage{algpseudocode}

\DeclareMathOperator{\EX}{\mathbb{E}}
\DeclareMathOperator{\tr}{tr}
\usepackage[colorlinks=true, pdfstartview=FitV, linkcolor=black, citecolor= black, urlcolor= black]{hyperref}
\usepackage{overcite}
\usepackage{footnpag}			      	

\PaperNumber{XX-XXX}

\begin{document}

\title{Diffusion Models for Generating Ballistic Spacecraft Trajectories}

\author{Tyler Presser\thanks{Ph.D. Student, Department of Astronautical Engineering, University of Southern California, tpresser@usc.edu},  
Agnimitra Dasgupta\thanks{Postdoctoral Research Associate, Department of Aerospace and Mechanical Engineering, University of Southern California, adasgupt@usc.edu },
Daniel Erwin\thanks{Professor, Department of Astronautical Engineering Department, University of Southern California, erwin@usc.edu},
\ and Assad Oberai\thanks{Hughes Professor and Professor of Aerospace and Mechanical Engineering, University of Southern California, aoberai@usc.edu}
}

\maketitle{}

\begin{abstract}
Generative modeling has drawn much attention in creative and scientific data generation tasks. Score-based Diffusion Models, a type of generative model that iteratively learns to denoise data, have shown state-of-the-art results on tasks such as image generation, multivariate time series forecasting, and robotic trajectory planning. Using score-based diffusion models, this work implements a novel generative framework\footnote{Training data and codes developed are publicly available at https://github.com/tpresser570/Lambert-Diffusion} to generate ballistic transfers from Earth to Mars. We further analyze the model's ability to learn the characteristics of the original dataset and its ability to produce transfers that follow the underlying dynamics. Ablation studies were conducted to determine how model performance varies with model size and trajectory temporal resolution. In addition, a performance benchmark is designed to assess the generative model's usefulness for trajectory design, conduct model performance comparisons, and lay the groundwork for evaluating different generative models for trajectory design beyond diffusion. The results of this analysis showcase several useful properties of diffusion models that, when taken together, can enable a future system for generative trajectory design powered by diffusion models.
\end{abstract}

\section{Introduction}
 Diffusion models are a type of generative model that have achieved state-of-the-art performance across creative and scientific domains. They have transformed the world of image generation and make up the backbone of many well-established models such as DALL-E 2\cite{Ramesh2022HierarchicalTI} and Stable Diffusion\cite{rombach2022highresolution}. Concerning trajectory design, diffusion models have shown promising results in robotics. Janner et al. propose combining diffusion models with reinforcement learning techniques to develop flexible trajectory planning strategies\cite{janner2022planning}. Other applications in robotics perform high fidelity robotic behavior generation\cite{chen2023offline}, enable visual robotic manipulation\cite{ryu2023diffusionedfs}, sample both task and motion planning under partial observability\cite{fang2023dimsam}, and can stitch together robotic actions into a coherent motion plan for offline Reinforcement Learning (RL)\cite{kim2024stitching}. 

Guffanti et al.\cite{guffanti2024transformers} describe AI-based trajectory design for spacecraft trajectories as an emerging research field, primarily categorized into two distinct segments. The first segment focuses on learning a representation of action policies, value functions, or reward models by applying either RL or Supervised Learning (SL) techniques. In contrast, the second segment employs learning-based elements to warm-start sequential optimization solvers. This study demonstrates the potential use of diffusion models in future trajectory design systems to produce viable trajectories that can be integrated into systems of the second category.

In this work, we implement a novel framework for trajectory generation built upon score-based diffusion models\cite{song2020generative}. The core idea is to train a diffusion model on a database of Earth-Mars transfers produced via Lambert's solution and generate new feasible trajectories. Score-based generative models\cite{song2020generative} are a class of diffusion models that utilize neural networks to approximate the score\cite{pmlr-v48-chwialkowski16} function of a target distribution. New samples are generated from the target distribution using annealed Langevin dynamics\cite{langevin}. The diffusion models utilized in this work are unconditional, which means that the generation process is unsupervised and does not take in user inputs to guide the model.

As a prototype trajectory design problem, the model is tasked with producing ballistic transfers from Earth to Mars within the departure window 01/01/2005-01/01/2006. The transfers are limited to prograde and short-way Lambert solutions, and only the planar two-body problem dynamics are considered in this preliminary study. Transfer Time Of Flight (\(\mathrm{TOF}\)) is limited to 120-270 days following the work done by Landau and Longuski\cite{landau} on human Mars transfers, and data are generated in 2-day increments. The training dataset is created using the Lambert solver provided by Izzo\cite{Izzo_2014} et al. through the \textit{Pykep}\footnote{https://esa.github.io/pykep/} Python library for astrodynamics. We generate more than 26,000 transfers from Earth-Mars and propagate them in Julia. The score-based model is trained to estimate the score of a given sample based on the sample's noise level, utilizing denoising score-matching\cite{song2020generative} techniques. However, the loss objective is uninformative regarding the model's ability to generate valid transfers. To test and evaluate the proposed framework for its usefulness in astrodynamics, we develop a new metric, the Defect Root-Mean-Square(RMS) Number (DRN), which allows the user to measure how well the generated trajectory obeys the underlying dynamics relative to the tolerances required for convergence in an optimization tool. In addition, we conduct several model ablation studies to assess model performance with varying temporal resolution and model sizes. 

Thus, the key contribution is the design, implementation, and testing of a novel probabilistic framework that uses diffusion models to generate new trajectories not included in the original training dataset while still capturing the original dataset's characteristic properties. In the style of previous work by Janner et al. \cite{janner2022planning}, we discuss several appealing properties of applying diffusion models to generate spacecraft trajectories that apply beyond the Earth-Mars transfer prototype problem. First, we demonstrate the ability of diffusion models to generate trajectories that retain key characteristics of the training dataset. Second, we showcase the diffusion model's ability to capture underlying dynamics with stable error under varying time horizons. This is a direct result of the model generating all trajectory states simultaneously rather than sequentially. Finally, we show that the framework's architecture enables the straightforward extension to complex tasks beyond the prototype problem.

\section{Background}
This section is intended to provide an overview of the key mechanisms and algorithms within diffusion models that enable the proposed framework. While the focus of this work is on a specific family of diffusion models known as score-based diffusion, there are several other well-known types of models, like the DDPM\cite{Diffusion} and DDIM\cite{song2022denoising}, that operate under the same set of underlying principles and can be shown to be different methods to parameterize the same process\cite{meng2022sdedit}. 

Song and Ermon first introduced score-based diffusion modeling in 2019\cite{song2020generative}. Since then, score-based models and their variants have been able to achieve state-of-the-art performance on image sample generation\cite{meng2022sdedit}, audio synthesis\cite{zhang2023survey}, and shape generation \cite{koo2024salad}. These models have also been well documented in several papers detailing processes for model construction\cite{song2020generative}, hyperparameter selection\cite{song2020improved}, and developing the mathematical framework used to design these models\cite{song2020improved}.  This section first describes the general process for generative modeling and the difference between unconditional and conditional tasks. We then detail the theory behind score-based generative modeling, beginning with a discussion of the score function and its role. Next, we show how Song and Ermon design a generative architecture to approximate the score function\cite{song2020generative} and employ annealed Langevin dynamics\cite{langevin} to generate new samples with an optimized score estimator. Finally, we briefly overview the trajectory models and tools for creating our dataset. 

\subsection{Generative Modeling}
Song \cite{yangsongGenerativeModeling} puts the objective of generative modeling as: given a dataset ${(\bm{x}_1,\bm{x}_2,...,\bm{x}_N)}$ where each point is drawn independently from a distribution \(p_{\mathrm{data}}(\bm{x})\), to fit a model to this distribution that allows us to sample new points from that distribution. Unconditional generation, which is the focus of this work, seeks to generate new data samples from \(p_{\mathrm{data}}(\bm{x})\). For diffusion models, this implies that the generation process is entirely unsupervised, and the model is tasked with denoising an initial random vector into any new sample that would be representative of the underlying data distribution. In generative machine learning research, unconditional generation is often used to test model performance on new foundational architectures or on specific prototype problems or datasets to validate a model's ability to learn to generate valid samples\cite{li2024return}. Diffusion models\cite{Diffusion}, Variational Autoencoders (VAEs)\cite{kingma2022autoencoding}, and Generative Adversarial Networks\cite{NIPS2014_5ca3e9b1} (GANs) were originally developed for unconditional generation. Unconditional generation is the focus of this work: we use it to validate the use of the Diffusion models to generate ballistic spacecraft trajectories. The new trajectories generated with the diffusion model are approximately within the training dataset's distribution \(p_{\mathrm{data}}(\bm{x})\). It is worth noting that conditional generative models, which are the subject of future work, utilize user-defined inputs to supervise the generative process and produce specific samples.

\subsection{The Score Function}
It is clear then that to build any generative model, one needs a method of representing any arbitrarily complex probability distribution. To solve this problem, Song and Ermon propose using neural networks to estimate the gradient of the log probability density function\cite{song2020generative}. This quantity is called the \textit{score}\cite{pmlr-v48-chwialkowski16} function. The score function of a distribution \(p_{\mathrm{data}}(\bm{x})\) is defined as $\nabla_x \log{(p_{\mathrm{data}}(\bm{x}))}$. The score is a vector field that points in the direction where the log data density grows the fastest at an input point. More intuitively, the vector field produced by the score function informs us of the direction of the steepest ascent over the probability density. A model used to represent the score function is called a score network, defined as:

\begin{equation}
\label{score_net}
s_{\bm{\theta}}(\bm{x}) \approx \nabla_x \log{(p_{\mathrm{data}}(\bm{x}))} \,, 
\end{equation}

where \(\bm{\theta}\) is the parameters of a neural network. 

\subsection{Score-Based Diffusion Models}

\subsubsection{Training}
Song and Ermon's key insight is that a neural network can be trained to estimate the score function directly from data using denoising score matching.\cite{denoise_sm} The immediate question that follows this notion is how to train a model that can accurately estimate the score function from a dataset where we do not know the score \textit{a priori}. From a high-level perspective, it is necessary that the optimal model would minimize the objective: 

\begin{equation}
\label{fisher}
\frac{1}{2}\EX_{p_{\mathrm{data}}(\bm{x})}[||s_{\bm{\theta}}-\nabla_{\bm{x}}\log{p_{\mathrm{data}}}(\bm{x})||^2_2] \,,
\end{equation}
where $|| \cdot||^2_2$ is used to represent the squared-$\mathcal{L}_2$ norm of a vector.  Equation~\eqref{fisher} resembles the Fisher divergence\cite{Johnson_2004}, which compares the squared $\mathcal{L}_2$ distance between the data's true score and the model predicted score. While intuitive, this objective is non-useful since we do not know the true score. To get around this problem, a family of methods called \textit{score matching}\cite{scorematch} has been developed to minimize Equation~\eqref{fisher} without needing to estimate \(p_{\mathrm{data}}(\bm{x})\) first. Following the procedures in previous works and again laid out by Song and Ermon\cite{song2020generative}, Equation~\eqref{fisher} can be rewritten as: 

\begin{equation}
\label{fisher_rewritten}
\EX_{p_{\mathrm{data}}(\bm{x})}[||\tr(\nabla_{\bm{x}}s_{\bm{\theta}}(\bm{x}))-\frac{1}{2}\nabla_{\bm{x}}\log{p_{\mathrm{data}}}(\bm{x})||^2_2]\,,
\end{equation}
where $\tr(\cdot)$ is the matrix trace operator. Although Equation~\eqref{fisher_rewritten} is only dependent on the neural network \(s_{\bm{\theta}}\), it is not scalable to high dimensional data since the operation within the trace is equivalent to computing the Jacobian of  \(s_{\bm{\theta}}\) which in practice is too large to compute efficiently. To completely circumvent $\tr(\nabla_{\bm{x}}s_{\bm{\theta}}(\bm{x}))$ Song and Ermon\cite{song2020generative} suggest adopting denoising score matching\cite{denoise_sm}. Denoinsing score matching starts by perturbing \(p_{\mathrm{data}}(\bm{x})\) by a sequence of noise scales $\{\sigma_i\}_{i=1}^L$ where \(\sigma_1 > \sigma_2 > ... > \sigma_L\) and \(\sigma^2\) is the variance of the Gaussian noise used to perturb \(\bm{x}\). Now, we define the Gaussian noise as \(p_{\sigma}(\Tilde{\bm{x}}|\bm{x}) = \mathcal{N}(\Tilde{\bm{x}}|\bm{x},\sigma^2\mathbf{I})\), where $\mathcal{N}$ represents the Gaussian Normal Distribution, and can produce the entire noise-perturbed distribution as: 

\begin{equation}
\label{integral}
p_{\sigma}(\Tilde{\bm{x}}) \triangleq \int p_{\sigma} (\Tilde{\bm{x}}|\bm{x}) p_{\mathrm{data}}(\bm{x}) d\bm{x}
\end{equation}

Using this information Song and Ermon\cite{song2020generative} propose estimating the score at each noise level \(p_{\sigma_i}(\bm{x})\) by training the neural net conditioned on \(\bm{x}\) and \(\sigma\), noted as \(s_{\bm{\theta}}(\bm{x},\sigma)\). This updated definition is now called the Noise Conditional Score Network (NCSN) and has an updated loss function:

\begin{equation}
\label{denoising_loss}
\mathcal{L}(\bm{\theta}) = \frac{1}{2}\sum^L_{i=1} \EX_{p_{\mathrm{data}}(\bm{x})}\EX_{p_{\sigma_i}(\Tilde{\bm{x}}|\bm{x})}[|| \sigma_i s_{\bm{\theta}}(\Tilde{\bm{x}},\sigma_i) + \frac{\Tilde{\bm{x}}-\bm{x}}{\sigma_i}||^2 _2]
\end{equation}

Using Equation~\eqref{denoising_loss}, all expectations can now be estimated using empirical averages based on the Monte Carlo (MC) approximation of the expectation. An important property of Equation~\eqref{denoising_loss} and the denoising score matching techniques employed here is that no specific neural network architecture is required. The only constraint the loss function places is that the model input and output dimensionality match, which can be satisfied with many different model types. This architectural flexibility is attractive because it allows the designer to design models around hardware constraints directly. 

\subsubsection{Sampling}
Sampling from the model is done through Annealed Langevin dynamics, which can produce samples from a probability density \(p_{\mathrm{data}}(\bm{x})\) using only the trained score function. The implementation of Langevin dynamics for generating samples is shown in Algorithm \ref{langevin}. Three parameters need to be precomputed according to the techniques proposed by Song and Ermon\cite{song2020improved}. They are the fixed step size \(\epsilon\), sampling steps $T$, a sample \((\bm{x}_0)\) from any prior distribution \(\pi(\bm{x})\)  and noise scales $\{\sigma_i\}_{i=1}^L$. The process is presented in Algorithm \ref{langevin}. and can be understood more simply as noisy gradient ascent\cite{song2020improved} on \(\log p_{\mathrm{data}}(\bm{x})\). 

\begin{algorithm}
\caption{Annealed Langevin dynamics \cite{langevin}}
\label{langevin}
\begin{algorithmic}[1]
\State \textbf{Input:} $s_{\bm{\theta}}$, $T$, $\epsilon$, $\{\sigma_i\}_{i=1}^L$
\State Initialize $\bm{x}_0$ such that $\{\bm{x}_0\}_i \sim \mathcal{U}(0, 1)$
\For{$i = 1$ to $L$}
    \State $\alpha_i = \frac{\epsilon\sigma_i ^2}{\sigma_L ^2}$
    \For{$j = 1$ to $T$}
        \State Sample $\bm{z} \sim \mathcal{N}(\bm{0}, I)$
        \State $\bm{x}_t = \bm{x}_{j-1} + \alpha_i s_{\bm{\theta}}(\bm{x}_{j-1}; \sigma_i) + \sqrt{2\alpha_i}z$
    \EndFor
    \State Set $\bm{x}_0 = \bm{x}_T$
\EndFor
\State Denoise $\bm{x}_T = \bm{x}_0 + \sigma_L^2 s_{\bm{\theta}}(\bm{x}_0) $
\State \textbf{Output:}  Realization $\bm{x}_T \sim \pi(\bm{x})$
\end{algorithmic}
\end{algorithm}

When reviewing Algorithm \ref{langevin}, the approach can be thought of more intuitively as repeatedly sampling from the noisy distributions \(p_{\sigma_1}(\bm{x}), p_{\sigma_2}(\bm{x}),  ..., p_{\sigma_L}(\bm{x}) \). We also define the step schedule as \(\alpha_i = \epsilon \sigma_i ^2/\sigma_L ^2 \) for the $i^{th}$ noise scale. A final denoising step, step 11 in Algorithm \ref{langevin}, then denoises samples from \(p_{\sigma_{1}}(\bm{\bm{x}})\). The samples from a given noise scale are then used to initialize the next iteration of Langevin dynamics for the next noise scale. 

\section{Dataset Generation and The Lambert Problem}
The prototype problem chosen to test the viability of Diffusion models to generate trajectories is ballistic Earth-Mars transfer design. The simplest version of this problem is deemed the orbital boundary value problem, also referred to as the Lambert problem\cite{Izzo_2014}. To solve the Lambert problem, the user only needs to define an initial and final position vector \([\bm{r}_i,\bm{r}_f]\) and the desired time of flight, \(\mathrm{TOF}\), to connect those two positions. There are many publicly available code repositories with efficient Lambert solvers that all seek to produce the set of two velocities \([\bm{v}_i,\bm{v}_f]\) to define the two-body orbit connecting \([\bm{r}_i,\bm{r}_f]\). This work utilizes the Lambert solver designed by Izzo\cite{Izzo_2014}, which is made publicly available through the ESA's \textit{Pykep} Python package. 

The launch window for this analysis was chosen to be 01/01/2005-01/01/2006. The flight time varies from 120-270 days in increments of 2 days. The time of flight range was chosen based on an analysis done by Landau and Longuski\cite{landau} on trajectories for human missions to Mars. The dataset includes only zero-revolution, prograde, short-way transfers. We also choose to omit any transfer with a transfer angle \(\theta\), that is within \(175^\circ < \theta < 185^\circ\) to avoid known instabilities in Lambert solvers near \(\theta=180^\circ\). With these conditions in place, this results in 26,019 Lambert problems to solve. Using only one CPU core on a 2023 Macbook Pro with M2 processor and 16 GB of RAM, all solutions are computed with the \textit{Pykep} solver in less than 1 second. 

With all of the necessary solutions computed, we then collect all the solved initial velocities, \(\bm{v}_i\), and individually propagate them in Julia using the DifferentialEquations.jl package with the available Tsitouras-Papakostas 8/7 Runge-Kutta method\cite{TSITOURAS2017226}, implemented as TsitPap8. We use a relative and absolute integration tolerance of  \(10^{-12}\). For propagation, we employ planar two-body dynamics whose acceleration is shown below in Equation \eqref{two_body_x}

\begin{equation}
\label{two_body_x}
\Ddot{\bm{r}} = -\frac{\mu}{|\bm{r}|^3} \cdot \hat{\bm{r}} ,
\end{equation}
 where \(\mu\) is the central body's gravitational parameter, the Sun, in this prototype problem. The full 3-D problem with \(z\) components of the position and velocity components included, are left for future work. The resulting dataset consists of 26,019 trajectories, of which we use \(90\%\) for training and reserve \(10\%\) for validation steps to ensure the score network is not overfitting. The dataset's memory size depends on the trajectories' desired temporal resolution or number of time steps recorded during propagation. This analysis uses 16, 64, 256, and 1024 time step trajectories where each trajectory timestamps are determined as a linear range \(t_{end}/n\) where \(n\) is the time resolution. This results in datasets that are respectively 32.3 MB, 75.7 MB, 319.7 MB, and 1.07 GB each. 
 
\begin{figure}[htbp]
	\centering\includegraphics[width=4in]{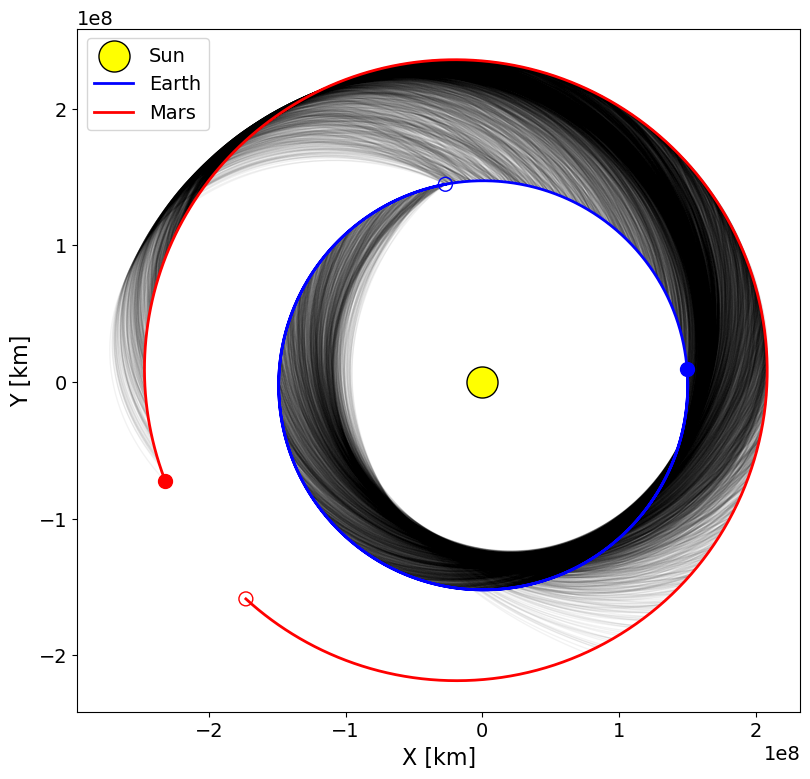}
	\caption{5000 randomly sampled trajectories from the Earth-Mars ballistic transfer training dataset.}
	\label{fig:dataset}
\end{figure}

 Figure \ref{fig:dataset} plots the propagated trajectory using planar two-body dynamics for initial conditions that are a concatenation of \(\bm{r}_1\) from the database set and its corresponding initial velocity provided by the Lambert solver for a specific \([\mathrm{TOF},\bm{r}_1,\bm{v}_1]\), for 5,000 randomly selected states from the dataset. Transparency is added to the trajectories to emphasize high-density regions in position space. The positions of Earth and Mars are plotted from 01/01/2005-09/28/2006, which encompasses the first launch date to the final arrival date. In Figure \ref{fig:dataset}, unfilled markers represent the positions of the planets at the first launch date, and filled markers represent their positions on the final arrival date. 

\section{Training and Evaluation Setup} 
Training generative networks, in general, is a compute-intensive process requiring several to hundreds of thousands of CPU and GPU hours\cite{song2020improved}. As such, generative AI research places great importance on training procedures to ensure efficient setup and training runs. After training, a generative network is typically evaluated by comparing model results against well-known benchmarks and tasks. For example, in image generation, this is often datasets like CIFAR-10\cite{cifar}, MNIST\cite{deng2012mnist}, and CelebA\cite{liu2015faceattributes} where the comparative benchmark is the Frechet Inception Distance or FID\cite{Seitzer2020FID}. As this work serves as the first application of diffusion models for spacecraft trajectory generation, this section outlines the training and evaluation setup used in generating transfers. We begin by detailing training procedures, including data preparation, hardware requirements, and model architecture ablation studies to be conducted. With training procedures outlined, we then discuss the proposed evaluation benchmarks that will be used to measure model performance. 

\subsection{Training the Model}
The model architecture used to generate all results in this work is the NCSNv2\cite{song2020improved} as implemented by Song and Ermon. NCSNv2 is based on the RefineNet\cite{lin2016refinenet} architecture, which is a variation of the well-known U-Net\cite{ronneberger2015unet} architecture for image generation and segmentation. In general, U-Nets are purpose-built to transform data through two paths: one contracting and one expanding path. This is the process of taking an initial input, contracting it to extract its most important features, and then transforming it into new information through expansion back to the original input shape, which gives these models their \say{U} shape.

With the score network architecture chosen, the next step is to prepare the data for training. For reference, key parameters of the dataset are summarized in Table \ref{tab:dataset}. Each trajectory is saved as a Python Pickle file, whose shape is \([5,n]\) where the five rows are \([t,x,y,v_x,v_y]\) and \(n\)  is determined by the dataset's temporal resolution. An important note here is that the NCSNv2 model is designed to work with images, not trajectories. The network expects a vector input shape \([c,w,l]\) where the dimensions represent the image's color channels, width, and height, respectively. By design, the width and height of the input image must be even numbers only, and input images are expected to be scaled between 0 and 1. 

\begin{table}[htbp]
	\fontsize{10}{10}\selectfont
    \caption{Training dataset key attributes.}
   \label{tab:dataset}
        \centering 
   \begin{tabular}{c | c | c | c } 
      \hline 
      Number of Trajectories & Launch Window & \(\mathrm{TOF}\) Range (days) & Distance Range (AU) \\
      \hline 
      26,019      & 01/01/2005 - 01/01/2006   & 120-270 & 0.955-2.665  \\
      \hline
   \end{tabular}
\end{table}

\begin{figure}[htb]
	\centering\includegraphics[width=5in]{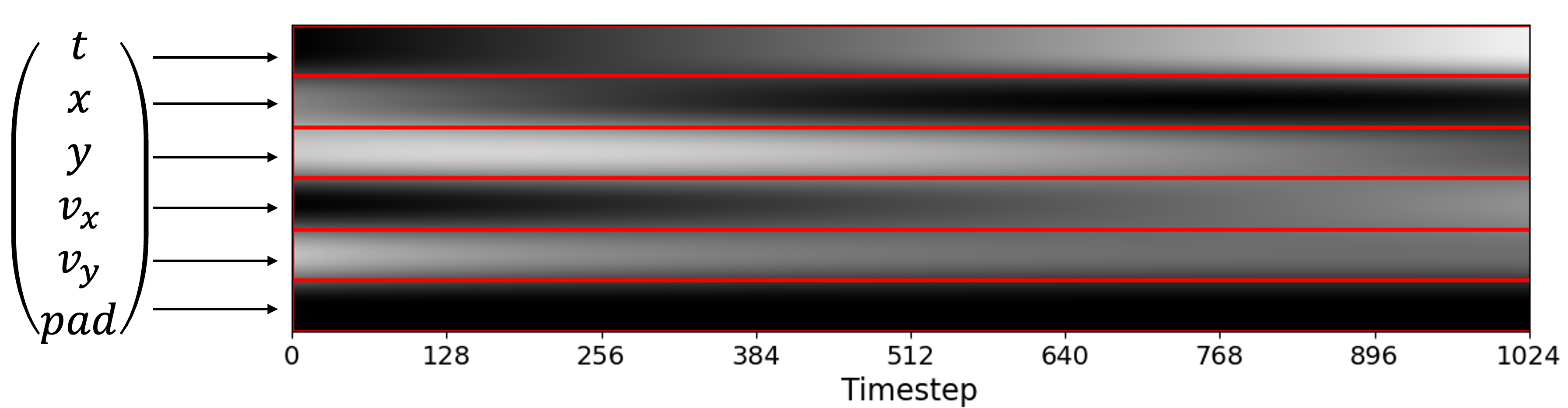}
	\caption{Example model input image and the corresponding state elements for each row in the image. Note, red boxes are added to differentiate between rows in the image.}
	\label{fig:image}
\end{figure}

We apply three simple operations to our dataset to reduce the number of changes needed to implement NCSNv2 for trajectory generation. First, we expand the dimensions of each vector to \([1,5,n]\), which transforms each trajectory into a single channel, grayscale image. Second, we expand the number of states to six with an additional row of padding zeros that do not affect the model output but do now meet the even dimension requirement. Finally, while creating the dataset, we record the minimum and maximum values of each of the 5 original state elements. Using these values, we individually scale each row in the input vector using min-max scaling to ensure all values are between zero and one. Figure \ref{fig:image} displays the result of this transformation from a trajectory vector, in this case, one with a 1024 time step resolution, to a single channel image that gets passed to the model. Note that the red boxes are added to this figure around each row to differentiate between rows and provide clarity between state elements.  

As mentioned, the output of NCSNv2 has the same shape as the input vector during training. After training is complete, we sample the model using annealed Langevin dynamics\cite{langevin}. The last step before starting training is to determine the model hyperparameters. For NCSNv2, Song and Ermon\cite{song2020improved} present several techniques for selecting model hyperparameters. Specifically they suggest best practices for selecting noise scales $\{\sigma_i\}_{i=1}^L$, number of Langevin dynamics steps \(T\), and step size \(\epsilon\). Determining model size is left to the reader and, in this work, is varied for an ablation study to determine changes in performance and efficiency with varying model sizes. The hardware used for each training run of NCSNv2 is held constant with  2 x A100 Nvidia GPUs and 64 x epyc-7513 AMD CPU cores. All training and sampling were conducted at the USC Center for Advanced Research Computing Discovery Cluster.\footnote{https://www.carc.usc.edu/}

\subsection{Evaluation}
During training, NCSNv2 is optimized by minimizing the objective defined in Equation \eqref{denoising_loss}. However, the final loss of the optimized model does not provide any insight into its ability to generate spacecraft trajectories. The defined loss function only informs the user that the model can estimate the score function for different noise levels. This work proposes three evaluation metrics to provide generative performance metrics relevant to mission design applications: a Lambert solver comparison test, the Defect RMS Number (DRN), and an Equivalent DRN that can be applied to arbitrary time step resolutions. When sampling each variation of NCSNv2, either with varying temporal resolution or model size, one thousand samples are always generated and used to compute the proposed evaluation metrics. All metrics are reported as scaled values to provide a sense of problem-specific performance for the prototype Earth-Mars transfer problem and to showcase the model's ability to complement existing optimization tools that utilize scaled inputs and constraints. For all evaluation metrics, model-generated positions are scaled by 1 AU, and all velocities are scaled by 30 km/s, which is developed as a rule of thumb for scaling heliocentric transfers from Earth by R\'e\cite{Parrish_2019}. The hardware used for sampling is held constant throughout this work and was set as: 2 x A100 Nvidia GPUs and 64 x epyc-7513 AMD CPU cores. 

\subsubsection{Lambert Solver Comparison}
The first evaluation metric directly compares the model-generated output to its real Lambert solution counterpart. The goal is to use the model output's initial state, final state, and generated \(\mathrm{TOF}\) as the inputs to the \textit{Pykep} Lambert solver. Using the solver's outputs, \([\bm{v}_i,\bm{v}_f]\), we can then directly compare the model's generated velocities at the initial and final states to determine how well the model learned to generate a trajectory that solves Lambert's problem. This process is outlined below and is repeated for the initial and final states.

\begin{enumerate}
  \item Extract \([\mathrm{TOF},\bm{r}_i,\bm{r}_f]\) from the model output states
  \item Use the \textit{Pykep} Lambert's solver\cite{Izzo_2014} to compute a solution and retrieve \([\bm{v}_i,\bm{v}_f]\)
  \item Compute the velocity magnitude difference between the model predicted velocity and the Lambert solution using $\Delta V = ||\bm{v}_{\mathrm{model}} - \bm{v}_{\mathrm{Lambert}}||$, at both times \(t_i\) and \(t_f\)
\end{enumerate}

\subsubsection{Defect RMS Number (DRN)}
While the Lambert solver comparison provides insight into the model's ability to solve Lambert's problem, it does not provide useful information about the model's ability to predict the entire trajectory while obeying the underlying two-body dynamics. To quantify the model's ability to plan and predict the entire trajectory, we propose a new metric: the Defect RMS Number (DRN), where RMS is the root-mean-square. The key concepts are derived from multiple-shooting optimization routines, which use the defect between nodes in a trajectory as the measure of convergence. Multiple shooting breaks a trajectory into \(N\) nodes, where each node has a state, velocity, and, in specific cases, control. We refer the reader to the work done by R\'e\cite{Parrish_2019} for further discussion on multiple-shooting implementations. For this work the DRN is computed by finding the defect along a trajectory at the midpoint between two adjacent nodes. The \(N-1\) defects at all the midpoints are then scaled by \(1\mathrm{AU}\) in position and \(30\mathrm{km/s}\) for velocity based on previous analysis by R\'e\cite{Parrish_2019}. This is based on requirements for optimization routines to have all variables scaled on the order of 1. Multiple shooting routines are typically considered to be converged when a constraint tolerance of \(10^{-10}\) is met\cite{Parrish_2019}. We design the DRN to be a single number representing how close a generated trajectory is to being considered converged. The process for computing the DRN is shown below:

\begin{enumerate}
  \item Compute all midpoint propagation times, \(t_{\mathrm{mid}} = \frac{t_{i+1}-t_i}{2}\)
  \item Propagate node \(i\) forward to time \(t_{mid}\), to get state \(\bm{X}_{forward}\)
  \item Propagate node \(i+1\) backward to time\(t_{mid}\), to get state \(\bm{X}_{backward}\)
  \item Compute the defect \(\bm{D}\) at node \(i\), using the state difference, \(\bm{D} = \bm{X}_{forward}-\bm{X}_{backward}\)
  \item Scale the defects vector positions by \(1  \mathrm{AU}\) and velocities by  \(30  \mathrm{km/s}\)
  \item Compute the $\mathrm{DRN}$ of $\bm{D}$ using the root-mean-square, $\mathrm{DRN} = \sqrt{\frac{1}{N-1} \sum_{i=1}^{N-1}\sum_{j=1}^4{\bm{D}^2_{ij}}          }  $, where \(N\) is the number of nodes in a trajectory
\end{enumerate}

\begin{figure}[htb]
	\centering\includegraphics[width=5in]{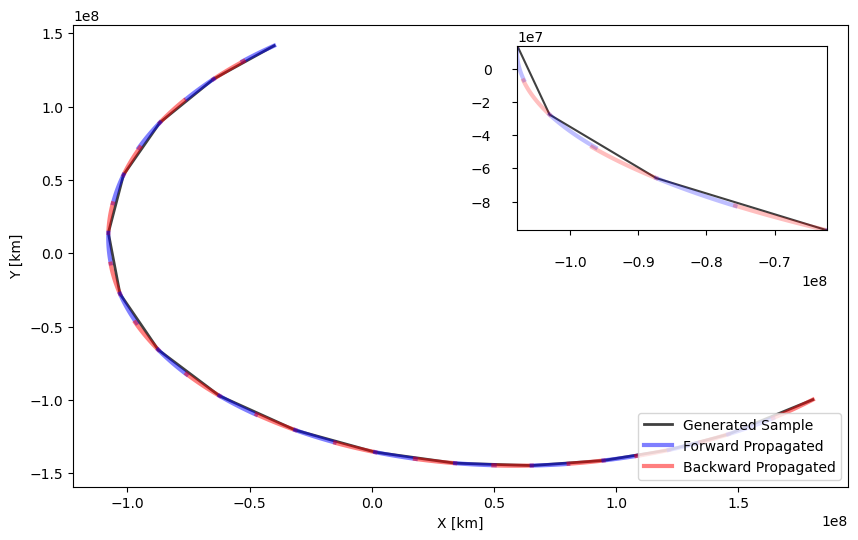}
	\caption{Forward-backward midpoint propagations for a 16-node trajectory. Defects computed at the midpoints are used to compute a trajectory's DRN. Note: the inset plot shows a zoomed-in view of three forward-backed propagations from the displayed trajectory, with transparency added to highlight overlapping sections.}
	\label{fig:multiple_shooting}
\end{figure}

Figure \ref{fig:multiple_shooting} visually displays the forward/backward propagation process for a 16-node generated trajectory. During model evaluation, we compute the DRN for all 1000 generated samples and record the mean and standard deviation. In addition to trajectory feasibility, the DRN can be used to compare how feasibility changes with time step resolution. As mentioned in the dataset generation section, we train NCSNv2 on varying time resolution trajectories: 16, 64, 256, and 1024 time steps. The DRN allows us to quantify the trade between the number of nodes in a trajectory and the feasibility of the generated trajectories.

\subsubsection{Equivalent DRN (EDRN)}
The Equivalent DRN (EDRN) extends the Defect RMS Number metric to attempt to produce an equivalent performance comparison between trajectories of different resolutions. As mentioned, the DRN gives the user a single number that determines the measure of convergence, or more simply feasibility, of a given trajectory. When discussing different temporal resolutions, comparing how feasible a trajectory is and if the model is learning to represent the underlying dynamics when the temporal resolution increases more accurately is useful. For trajectories with high temporal resolution, say 1024 time steps, the propagation time used to compute the defect is significantly smaller. These shorter propagation times result in smaller defects. Although there are more defect computations in higher resolution trajectories, it is not apparent that using the DRN alone provides insight into the model's ability to learn to predict the underlying dynamics more accurately. 

\begin{figure}[htb]
	\centering\includegraphics[width=6in]{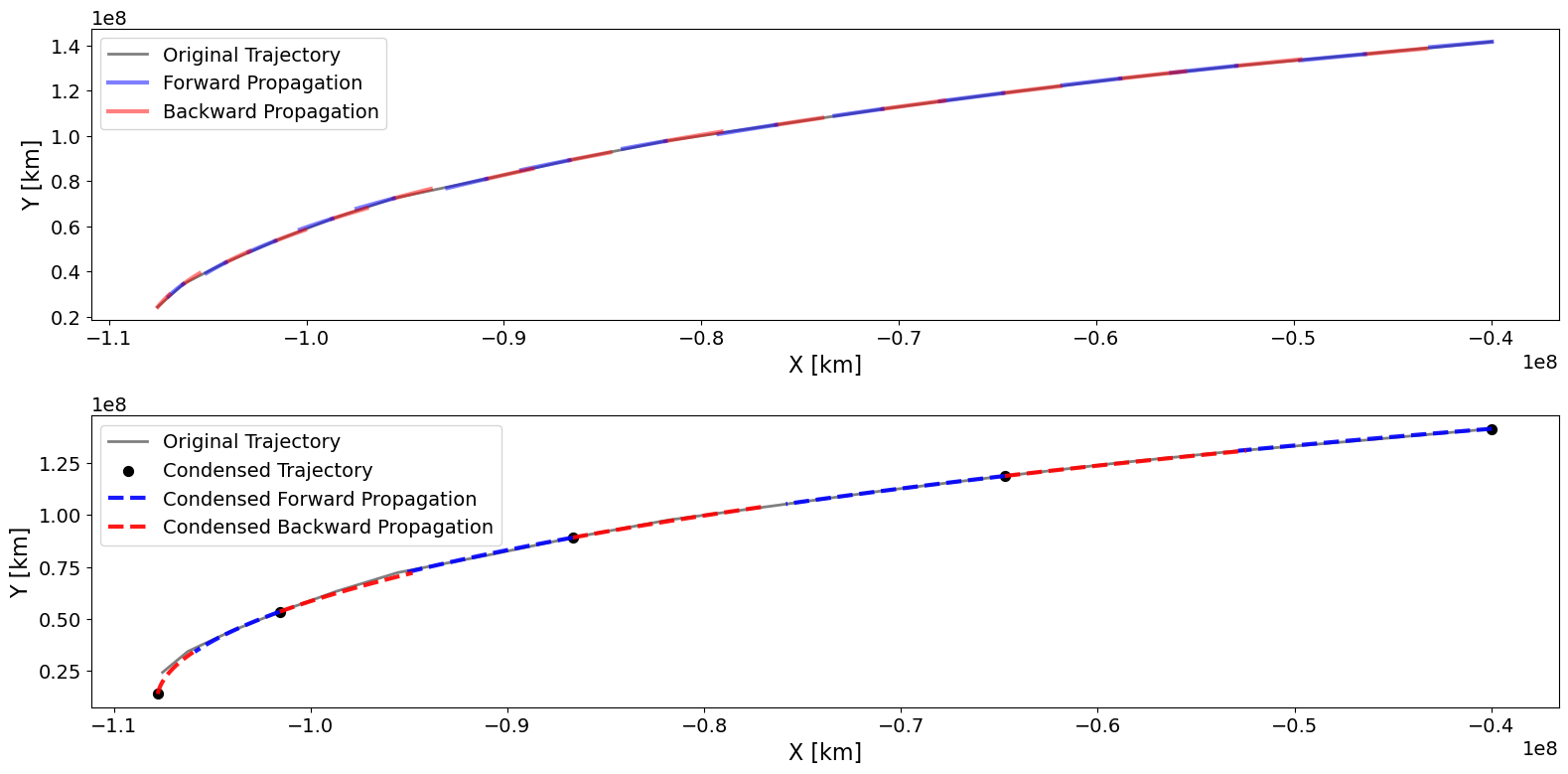}
	\caption{Condensation of a sixty-four node trajectory into sixteen nodes for computing the EDRN. Note that only a portion of the trajectory is shown.}
	\label{fig:windows_multiple_shooting}
\end{figure}

To this end, the EDRN metric provides direct insight into the relationship between temporal resolution and NCSNv2's ability to model the two-body dynamics. EDRN is computed to generate an equivalent comparison to the diffusion model's sixteen time step resolution version. For all temporal resolutions, computing the EDRN is done by collecting the initial and final nodes in the generated sample and then selecting fourteen time steps evenly spaced temporally between the two terminal states. With the samples reduced to sixteen time step resolution we then use the procedure outlined above to compute the DRN score for all 1000 samples to produce the EDRN. The process of condensing a sixty-four-resolution trajectory into a sixteen-resolution trajectory is shown in Figure \ref{fig:windows_multiple_shooting}.

\section{Results} 

This section discusses the proposed diffusion model's ability to generate ballistic transfers from Earth-Mars for the set analysis window. The core focus of our experiments is to determine the viability of diffusion models for use as trajectory design tools. Our goal is to evaluate the generated trajectory samples to determine if they: 1. Accurately capture key characteristics of the training dataset and 2. Obey dynamics constraints to generate feasible trajectories. We begin by discussing the model's ability to meet these two goals. We then perform two ablation studies that assess how model performance changes with model size. After determining the optimal model size, we vary the temporal resolution of the model inputs and output trajectories. Finally, we outline the most appealing properties of applying Diffusion models to build future trajectory design systems.

\subsection{Numerical Results}
\begin{figure}[h!]
	\centering\includegraphics[width=5.5in]{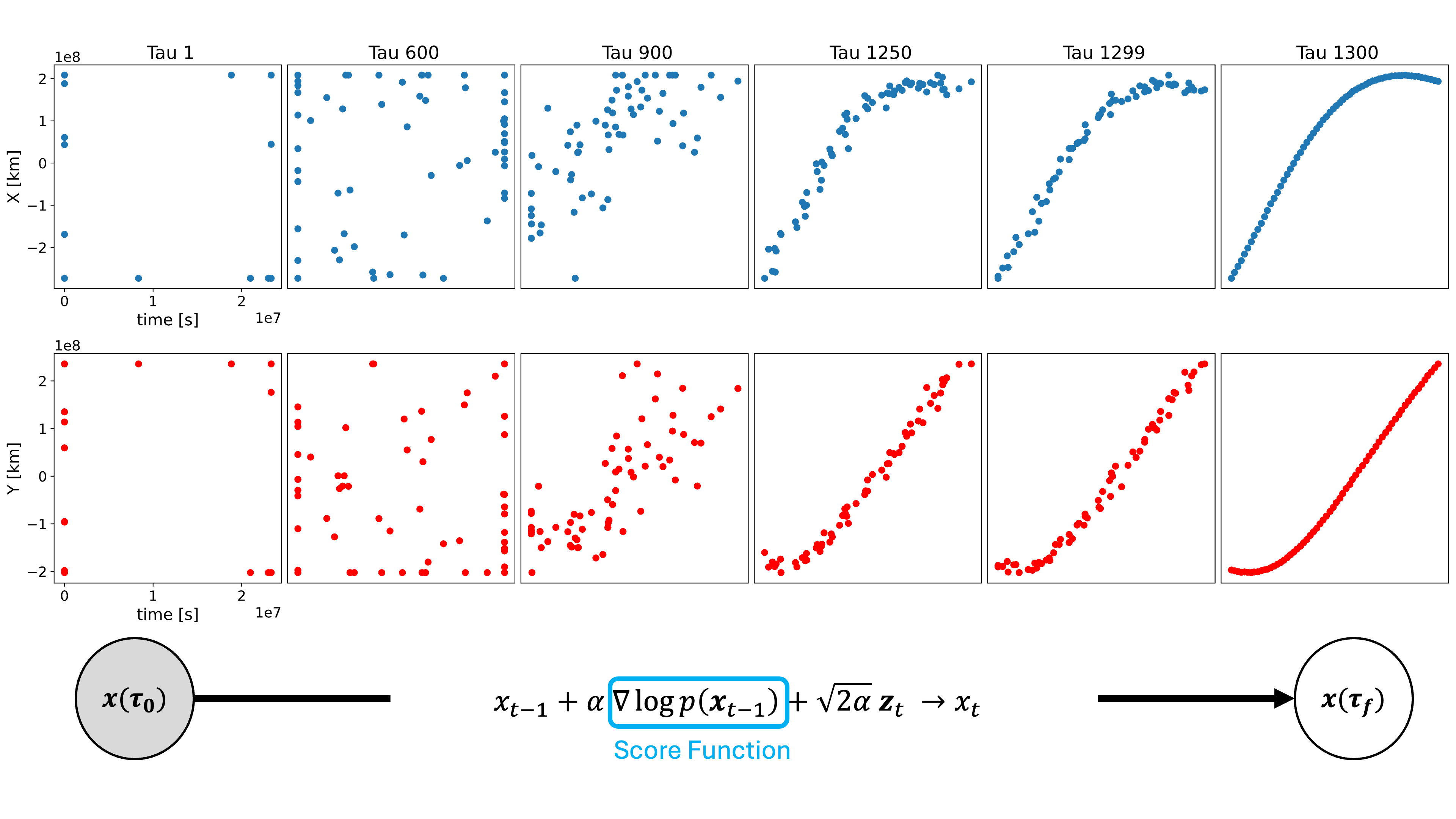}
	\caption{Trajectory generation process through iterative denoinsing shown for \(x\) and \(y\) state components. Note that $\tau$ here refers to the denoising step number.}
	\label{fig:denoise}
\end{figure}

The base model used to assess the outlined experimental goals was trained using 64 node trajectories and a model size indicated by model serial number 3 in Table \ref{tab:size_trade}. The training was completed in 8.6 hours, and during sampling, the model is capable of generating a new trajectory every 0.06 seconds on the previously described hardware setup in the Evaluation subsection. To illustrate the iterative denoising process, Figure \ref{fig:denoise} displays the model output over various denoising steps for the \(x\) and \(y\) position components of the trajectory state vector. In addition to model output we also display the demonising update step from Algorithm \ref{langevin}, and highlight the score function that the model is trained to estimate.

\begin{figure}[htb]
	\centering\includegraphics[width=6in]{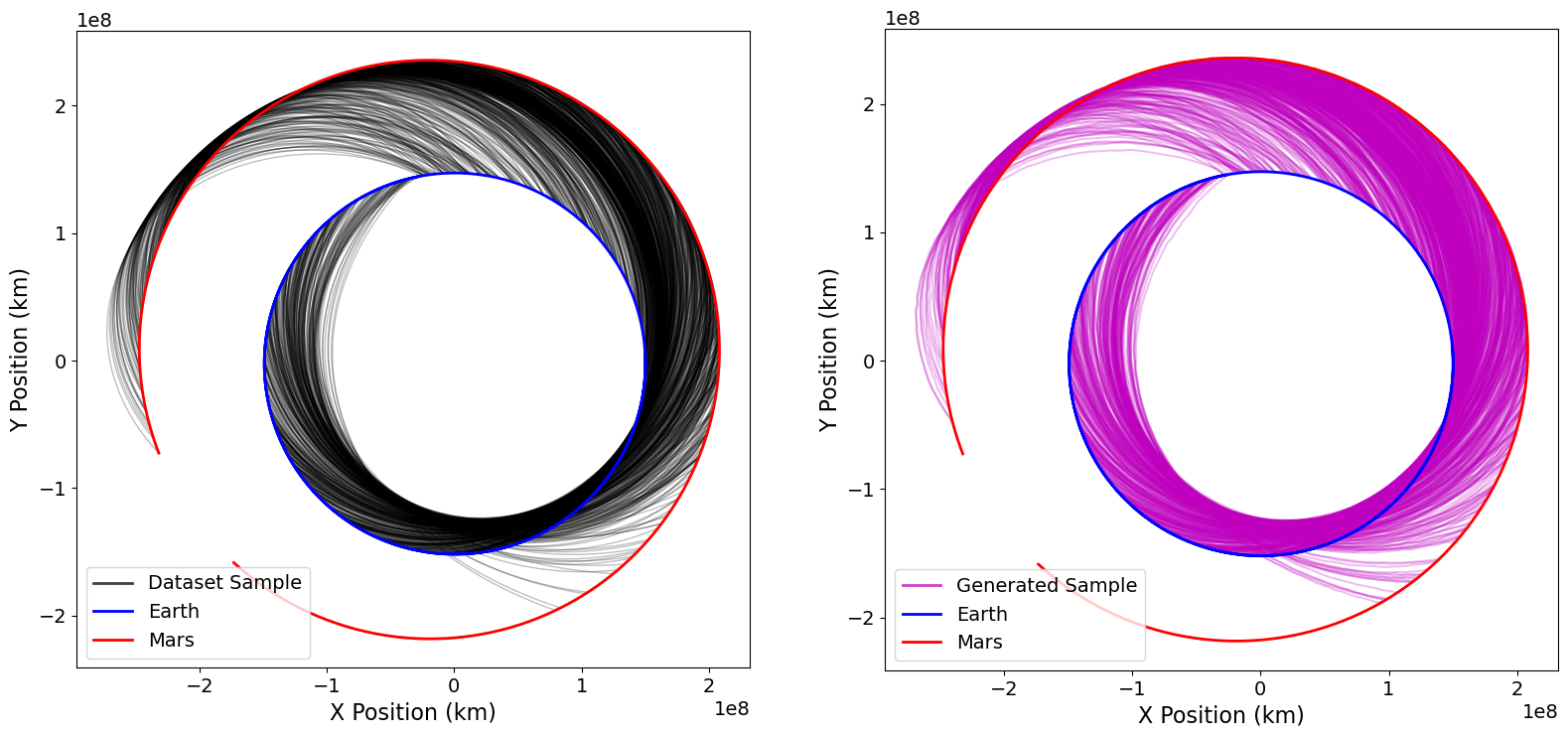}
	\caption{A side-by-side comparison of 1000 randomly selected trajectories from the training dataset and 1000 samples generated directly from the diffusion model.}
	\label{fig:earth_mars}
\end{figure}

\begin{figure}[h!tb]
	\centering\includegraphics[width=6in]{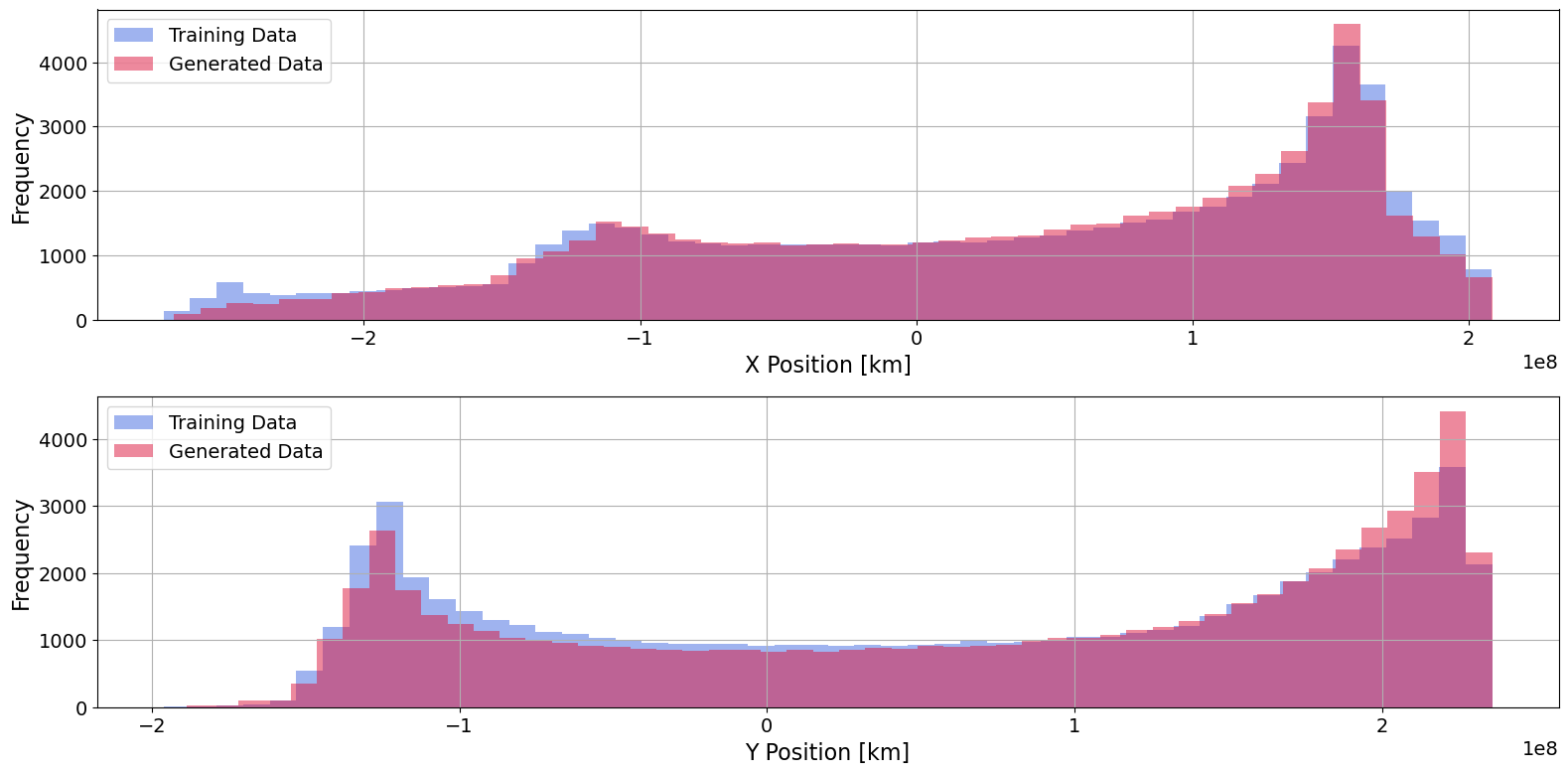}
	\caption{Histograms of position frequency for both \(x\) and \(y\) with generated values superposed on top of original dataset values for 1000 samples.}
	\label{fig:histogram}
\end{figure}

Regarding our first goal of capturing key characteristics of the training dataset, we can first visually inspect the model-generated trajectories. The model is designed to generate new samples that fit within and represent the training dataset's distribution. Figure \ref{fig:earth_mars} we plot a side-by-side comparison between 1000 randomly sampled trajectories from the training dataset on the left and 1000 model-generated new trajectories on the right. To illustrate high-density regions, we add transparency to the trajectories. As a result, darker regions of the plots represent high-density regions, and lighter trajectories represent relative outliers in the datasets. It is clear from visual inspection that the model could accurately produce samples from the original training data distribution. To better assess the difference between the two distributions, we can plot histograms of both the \(x\) and \(y\) positions. Figure \ref{fig:histogram} shows these histograms with the model-generated values overlaid on the values from the original dataset.

To quantify the base model performance, all outlined evaluation metrics for the 1000 model generated samples are shown below in Table \ref{tab:base_performance}. Note that the mean scaled difference between the model-generated velocities and the velocities produced by the Lambert solver are shown separately for \(\bm{v}_i\) and \(\bm{v}_f\). Recall that the results are computed according to the procedure outlined in the subsection on Evaluation and scaled to provide a sense of relative performance for the Earth-Mars transfer problem. 

\begin{table}[htbp]
	\fontsize{10}{10}\selectfont
    \caption{Base model key performance metrics. Note, \(\sigma\) is used to denote statistical standard deviation.}
   \label{tab:base_performance}
        \centering 
   \begin{tabular}{c | c | c | c } 
      \hline 
      Lambert \(\bm{v}_i\) Mean Difference  & Lambert \(\bm{v}_f\) Mean Difference & Mean DRN & DRN \(\sigma\)\\
      \hline 
      3.69e-3      & 3.02e-3   & 2.83e-3 & 4.88e-4  \\
      \hline
   \end{tabular}
\end{table}

Concerning our second goal of capturing the underlying two-body dynamics, the trajectories produced by the diffusion model have errors on the order of \(10^{-3}\) across all metrics. Using the mean DRN value, these represent errors in position and velocity that are 0.28\% of their scaling values. These errors translate to unscaled position and velocity magnitudes of \(10^{-3} \mathrm{AU}\) and \(100 
 \mathrm{m/s}\). When comparing these results to other applications of diffusion models, we find that the largest source of error in this work is the FP16 precision used in NCSNv2. To be accurate within a kilometer of position, the model output would need to be accurate to \(10^{-8}\) after the min-max scaling is applied to the original dataset. This type of accuracy would be nearly impossible without higher floating point precision data.  

The previously mentioned ablation studies are designed to determine the optimal trajectory temporal resolution and model size to maximize performance. We perform this study using the 16-time step resolution and gradually increase the model size. We refer the reader to the techniques developed by Song and Ermon\cite{song2020improved} for determining appropriate model sizes for NCSNv2. The control model size used in this study is model serial number 3. The results of the model size study are shown in Table \ref{tab:size_trade}. In Table \ref{tab:size_trade}, we use \(\sigma\) to abbreviate standard deviation. We note that the 16 time step model was used for this trade study because it requires the shortest training time. In principle, this study could be run for any temporal resolution.

\begin{table}[htbp]
	\fontsize{10}{10}\selectfont
    \caption{Model size comparison on 16 time step trajectories}
   \label{tab:size_trade}
        \centering 
   \begin{tabular}{c | c | c | c | c } 
      \hline 
      Model Serial Number  & Model Parameters & Model Size (MB) & Mean DRN & DRN \(\sigma\)\\
      \hline 
      1       & 117078   & 2.0 & 3.26e-3 & 8.90e-4 \\
      2      & 466107   & 7.7 & 3.03e-3 & 8.60e-4 \\
      3      & 1859211  & 30 & 2.96e-3 & 8.50e-4 \\
      4      & 7427115  & 119 & 2.98e-3 & 8.52e-4 \\
      \hline
   \end{tabular}
\end{table}

The most prominent insight from the size trade study is that for the 16-time step model, we can decrease the number of parameters from the base model, serial number 3, by 93.7\% while only losing 2.4\% in performance. For this study, we choose to continue with the model serial number 3 size since it is the most performant. However, for memory-constrained systems, only a small drop in performance must be sacrificed for large memory savings. 

The second trade study assesses model performance with varying temporal resolution. Again NCSNv2 is designed to output vectors that are the same shape as its inputs. Following the results of the first trade study, we use a constant model size parameter of 32 for all models trained in the temporal resolution study. Table \ref{tab:length} shows the training time and DRN performance for all tested temporal resolutions. From this trade study, we find that while the highest resolution performs the best both on mean DRN and has the smallest standard deviation, the additional time needed for training is an order of magnitude higher than all other lengths tested. 

To determine if the higher resolution models can more accurately obey two-body dynamics, we employ the Equivalent DRN (EDRN) outlined in the Evaluation subsection. The EDRN results for all tested temporal resolutions are shown in Table \ref{tab:size_trade}. On both DRN and EDRN the 64 time step model has the best performance. The model size trade study also informs us that without memory constraints, a model size based on model serial number 3 provides the highest level of performance.

\begin{table}[htbp]
	\fontsize{10}{10}\selectfont
    \caption{Comparison of model performances with varying output trajectory length.}
   \label{tab:length}
        \centering 
   \begin{tabular}{c | c | c | c | c | c  } 
      \hline 
      Resolution & Training Time (hrs) & Mean DRN &  DRN \(\sigma\) & Mean EDRN & EDRN \(\sigma\)

      \\
      \hline 
      16   & 8.5 &2.96e-3 &  8.50e-4 & N/A & N/A\\
      64   & 8.6 & 2.83e-3 & 4.88e-4 & 2.95e-3 & 6.84e-4 \\
      256  & 9.0 &2.85e-3  & 3.81e-4 & 3.00e-3 & 6.47e-4 \\
      1024 & 19.9 & 2.81e-3  & 3.23e-4 & 3.00e-3 & 6.25e-4\\
      \hline
   \end{tabular}
\end{table}

\subsection{Key Properties of Diffusion Models for Trajectory Design Systems}
\subsubsection{Ability to design feasible trajectories}
This analysis demonstrates a diffusion model's ability to generate new ballistic trajectories from a learned distribution. Our prototype problem tasked the model with generating ballistic trajectories from Earth to Mars. We evaluate the feasibility of these generated trajectories based on their difference in velocity magnitude at both initial and final states with the equivalent Lambert solution. In addition, we use the midpoint defect between nodes to measure how close to \say{real} the entire generated trajectory is. The generated trajectories' initial and final velocity error averages approximately 100 m/s from the velocities produced for the same initial and final positions, and \(\mathrm{TOF}\), using a Lambert solver. The developed DRN metric evaluates the midpoint defects and is average on the order of \(10^{-3}\) for all model variations. While this error level is too high to be considered converged, the trajectories generated by the diffusion model would make viable initial guesses for optimization routines like multiple shooting. These routines are normally considered to be converged once all constraints have been met to within a tolerance of \(10^{-10}\). The simplest path forward for incorporating diffusion models into trajectory design systems is to use them directly as automated initial guess tools. A future conditional model could produce many viable initial guesses for the same user-defined inputs in parallel. Since sampling time depends only on model size and time step resolution, hundreds of initial guesses could be produced in seconds, and the user could select specific generated samples with desirable characteristics. This allows the mission designer to trade objectives like time of flight or fuel mass in near real-time.

\subsubsection{Stable Error}
\begin{figure}[htb]
	\centering\includegraphics[width=6in]{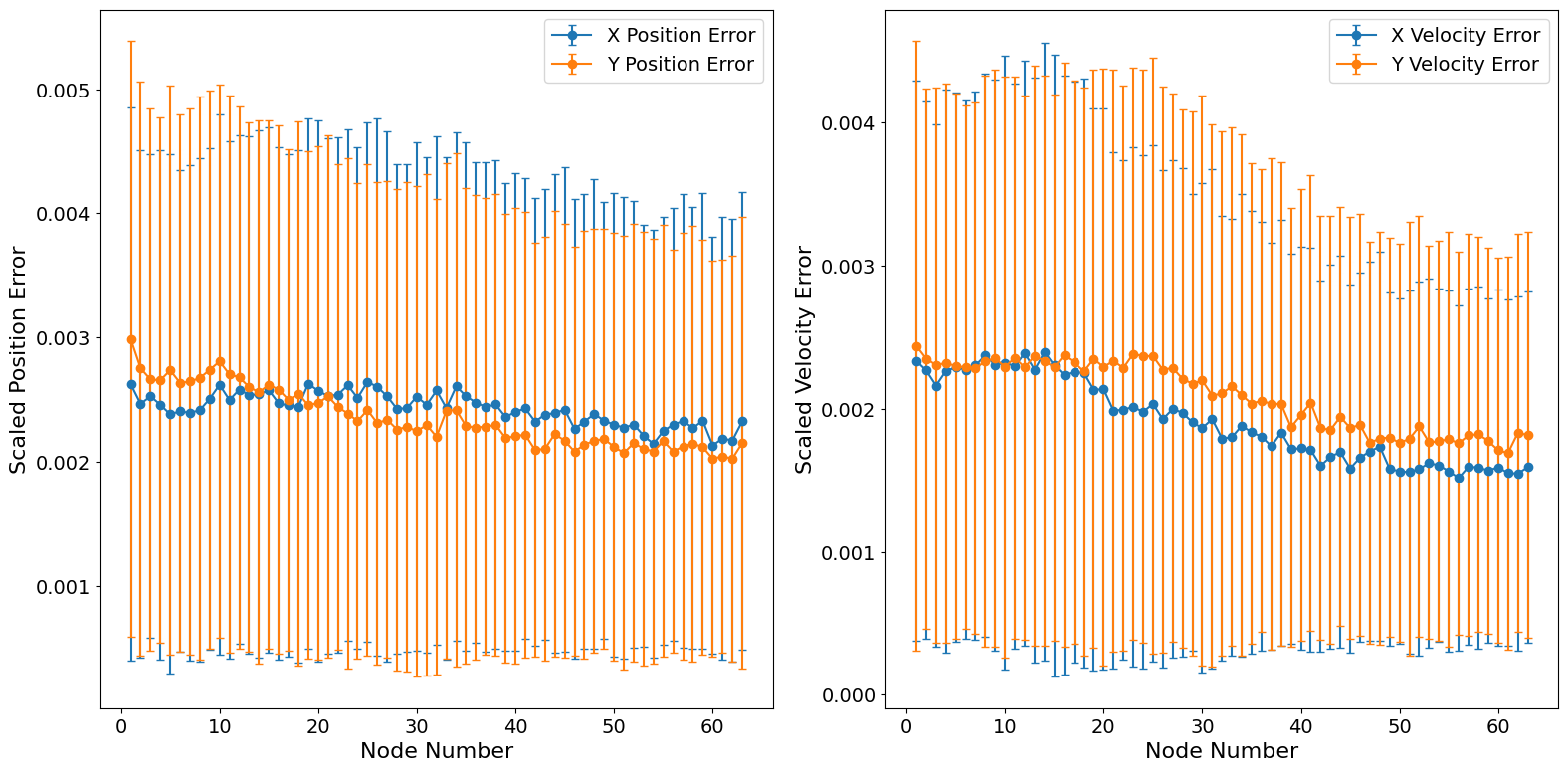}
	\caption{Mean position and velocity defects across nodes from 1000 samples generated with the base NCSNv2 model.}
	\label{fig:shooting_error}
\end{figure}

An important property of the diffusion model is a consistent error across time steps. When computing the DRN for a batch of 1000 trajectories, we can record the defects in position and velocity at each node and average them over the entire batch to show that the defects remain nearly constant in all state elements over the entire trajectory. For the base model output 64 node trajectory, the average error and standard deviation at each node are plotted in Figure \ref{fig:shooting_error}. From Figure \ref{fig:shooting_error}, we see that the average error across all 1000 generated trajectories remains nearly constant regardless of the propagation time as represented by the node. This appealing property of diffusion models is directly related to the generation process. The model predicts all states in a trajectory simultaneously rather than sequentially. Since the model is tasked with estimating the score function, its objective, in the simplest sense, is to make the trajectory increasingly similar to a sample from the underlying data as noise is removed. The score is a vector with the same size as the trajectory and is used to adjust all states simultaneously. This results in an even distribution of error across the trajectory without bias toward any specific time along the trajectory. 

\subsubsection{Straightforward application to complex tasks}
The only requirement for training set by the model architecture is that the input and output dimensions must be the same shape. Adding additional input rows or channels to the image vector is straightforward to include 3-dimensional data or control inputs. To generate specific trajectories based on user inputs, much work has already been done on conditional generation with diffusion models\cite{zhang2023survey}. The common theme in these works is the addition of a context vector, which in this case would be the problem parameters such as initial and final conditions. The context vector combined with guidance techniques such as cross-attention or conditional instance normalization enables diffusion models to generate conditional samples from the training data distribution. Another aspect of the presented diffusion model is that we include time as one of the model outputs. While we choose constant time steps between nodes, the user may use adaptive steps to incorporate maneuvers such as gravity assists. Finally, multiple diffusion models could be used in sequence, each using the output from the model before it as guidance to increase fidelity or add specific features to the trajectory, like impulsive maneuvers.

\section{Conclusion}
This work serves as a proof of concept for assessing the viability of diffusion models to design spacecraft trajectories. We demonstrate that a diffusion model can learn to generate ballistic transfers from Earth to Mars. These transfers have initial and final velocities within approximately 100 m/s of the velocity magnitude error of the equivalent Lambert solution. We propose a new metric, the Defect RMS Number (DRN), to evaluate the feasibility of generated trajectories. In addition, we perform a trade study of model size and trajectory temporal resolution to determine the optimal model and input size for the prototype problem. We discuss several appealing properties of diffusion models for trajectory design, namely, their ability to model the characteristics of the training dataset and produce solutions with stable error regardless of the time of flight. 

Future work will focus on further developing conditional diffusion models to generate trajectories based on user-defined problem parameters. We also plan to tackle more challenging problems with conditional architectures by moving to multi-body dynamics and incorporating control with low-thrust transfers. The simplest and potentially most effective to incorporate the core concept of diffusion models in trajectory design systems will be as automated initial guess tools for traditional optimization methods.

The greatest challenge in incorporating diffusion models as standalone trajectory design tools is increasing the fidelity of generated trajectories. However, much work can still be done to improve diffusion model performance. Two such lines of work include using state-of-the-art fundamental models like DDIM\cite{song2022denoising} and designing a new generation process that uses a cascade of models to gradually increase sample fidelity. This is similar to the process that DALLE-2\cite{Ramesh2022HierarchicalTI} uses to generate images, using the output from one mode to see the next and increase image resolution gradually with several models.

\section{Acknowledgment}
The authors of this paper would like to thank Dr. Nathan R\'e and Dr. Jeffrey Parker for contributing to brainstorming for both prototype problems and model ablation studies, as well as helping to design the evaluation metrics used to test model performance. This work was funded by an Advanced Space LLC Research Fellowship. The authors also acknowledge the Center for Advanced Research Computing (CARC, carc.usc.edu) at the University of Southern California for providing computing resources that have contributed to the research results reported
within this publication.

\bibliographystyle{AAS_publication}   
\bibliography{references}   

\end{document}